\documentclass[conference]{IEEEtran}
\IEEEoverridecommandlockouts
\usepackage{cite}
\usepackage{amsmath,amssymb,amsfonts}
\usepackage{algorithmic}
\usepackage{graphicx}
\usepackage{textcomp}
\usepackage{xcolor}
\usepackage{bm}
\usepackage{hyperref}
\usepackage[round]{natbib}
\usepackage{graphicx} 
\usepackage{booktabs}
\usepackage{url}
\usepackage{float}
\usepackage{threeparttable}
\usepackage{booktabs}
\usepackage{array}

\def\BibTeX{{\rm B\kern-.05em{\sc i\kern-.025em b}\kern-.08em
    T\kern-.1667em\lower.7ex\hbox{E}\kern-.125emX}}
\begin{document}

\title{On the Tunability of Random Survival Forests Model for Predictive Maintenance}

\author{\IEEEauthorblockN{Yigitcan Yardimci}
\IEEEauthorblockA{\textit{Department of Statistics} \\
\textit{Eskisehir Technical University}\\
Eskisehir, Turkiye \\
yigitcanyardimcii@gmail.com}
\and
\IEEEauthorblockN{Mustafa Cavus}
\IEEEauthorblockA{\textit{Department of Statistics} \\
\textit{Eskisehir Technical University}\\
Eskisehir, Turkiye \\
mustafacavus@eskisehir.edu.tr}
}

\maketitle

\begin{abstract}
This paper investigates the tunability of the Random Survival Forest (RSF) model in predictive maintenance, where accurate time-to-failure estimation is crucial. Although RSF is widely used due to its flexibility and ability to handle censored data, its performance is sensitive to hyperparameter configurations. However, systematic evaluations of RSF tunability remain limited, especially in predictive maintenance contexts. We introduce a three-level framework to quantify tunability: (1) a model-level metric measuring overall performance gain from tuning, (2) a hyperparameter-level metric assessing individual contributions, and (3) identification of optimal tuning ranges. These metrics are evaluated across multiple datasets using survival-specific criteria: the \textit{C-index} for discrimination and the \textit{Brier score} for calibration. Experiments on four CMAPSS dataset subsets, simulating aircraft engine degradation, reveal that hyperparameter tuning consistently improves model performance. On average, the \textit{C-index} increased by 0.0547, while the \textit{Brier score} decreased by 0.0199. These gains were consistent across all subsets. Moreover, \texttt{ntree} and \texttt{mtry} showed the highest average tunability, while \texttt{nodesize} offered stable improvements within the range of 10 to 30. In contrast, \texttt{splitrule} demonstrated negative tunability on average, indicating that improper tuning may reduce model performance. Our findings emphasize the practical importance of hyperparameter tuning in survival models and provide actionable insights for optimizing RSF in real-world predictive maintenance applications.
\end{abstract}


\section{Introduction}

In machine learning, hyperparameters are essential elements that define the behavior and capacity of a model. Selecting appropriate hyperparameter values is critical for optimizing predictive performance, as poor configurations may result in overfitting, underfitting, or unreliable generalization and even predictive multiplicity \citep{cavus_et_al_2025} to unseen data. Hyperparameters often require empirical or data-centric tuning to achieve optimal generalization \citep{rijn_et_al_2018,probst2019tunability}. Consequently, hyperparameter tuning is key to developing robust and effective predictive models. This paper explores the impact of hyperparameter optimization within a real-world application domain in the context of predictive maintenance.

Predictive maintenance is a key industrial application where data-driven modeling provides tangible benefits. In this context, accurate predictions can lead to more efficient maintenance planning, while inaccurate ones may result in serious operational disruptions. However, inaccurate predictions may lead to unnecessary interventions or, conversely, to the failure of critical components, emphasizing the need for highly accurate and interpretable models \citep{vollert2021challenges}.

To support such high-stakes decisions, survival models must not only be accurate but also capable of handling censored data, unlike regression and classification models, which are common in real-world monitoring scenarios. Survival analysis emerges as a particularly suitable framework in this regard, as it accounts for censored observations—cases where the failure event has not occurred within the observation period. This makes it especially valuable for predictive maintenance tasks, where understanding time-to-failure is critical. Recent research has also demonstrated that appropriate hyperparameter optimization can substantially improve model calibration and discrimination across survival datasets \citep{baskay2025censoring,qin2022survival}. Among survival models, the Random Survival Forest (RSF) \citep{ishwaran2008random} stands out as a non-parametric, ensemble-based method that extends Random Forests \citep{breiman2001random} to time-to-event data. Its flexibility and robustness make RSF a widely adopted model across predictive maintenance applications.

However, the performance of RSF is highly sensitive to its hyperparameter configuration. In previous work, the performance of RSF was shown to vary significantly depending on the configuration of its key hyperparameters such as \texttt{nodesize}, \texttt{nodedepth}, and \texttt{mtry}, highlighting the need for systematic tuning strategies \citep{moradmand2021, dauda2022optimal}. Despite its widespread use, a systematic evaluation of the tunability—the extent to which performance can be improved by tuning hyperparameters—of RSF remains largely unexplored, especially in the predictive maintenance context. Although many software packages offer default hyperparameter values, these defaults may not be optimal for all datasets. In practice, users often rely on prior knowledge, trial-and-error, or empirical search to configure these values appropriately. 

This paper aims to contribute to this challenge by assessing the tunability of RSF across multiple datasets by proposing measurements for the tunability of survival models, similar to the work of \citet{probst2019tunability} conducted on the binary classification task. We systematically evaluate how tuning individual and multiple hyperparameters affects predictive performance, using established metrics \textit{C-index} and the \textit{Brier score}. The goal is to provide insights into the importance of hyperparameter optimization, to identify which parameters contribute most significantly to model improvement, and to determine the optimal searching range for the hyperparameters of the RSF model.

\section{Methodology}

Let $\mathcal{D} = \{(\mathbf{x}_i, T_i, \delta_i)\}_{i=1}^n$ denote a dataset with $n$ independent observations, where $\mathbf{x}_i = [x_{i1}, x_{i2}, \ldots, x_{ip}] \in \mathcal{X}$ is the predictor vector for the $i$-th observation, $T_i \in \mathbb{R}^{+}$ is the observed time-to-event, either failure or censoring, and $\delta_i \in \{0,1\}$ is the event indicator, with $\delta_i = 1$ indicating that the event (e.g., failure) was observed, and $\delta_i = 0$ indicating a censored observation. We assume that these triplets $(\mathbf{x}_i, T_i, \delta_i)$ are independently and identically distributed (i.i.d.) samples drawn from an unknown joint distribution $\mathcal{P}(\mathbf{X}, T, \delta)$, where capital letters represent the corresponding random variables. 

The goal is to learn a survival model $f^*: \mathcal{X} \rightarrow \mathcal{S}$ from a family of models $\mathcal{F} = \{f_{\bm{\theta}}\}$, parameterized by $\bm{\theta} = (\theta_1, \theta_2, \ldots, \theta_k)$ within a hyperparameter space $\bm{\Theta}$. The optimal model $f^*$ minimizes the expected value of a survival-specific loss function $L$:

\begin{equation}
    f^* = \underset{f_{\bm{\theta}} \in \mathcal{F}}{\arg\min}\, \mathbb{E}_{(\mathbf{X}, T, \delta) \sim \mathcal{P}} \left[ L(T, \delta, f_{\bm{\theta}}(\mathbf{X})) \right],
\end{equation}

\noindent where $L: \mathbb{R}^{+} \times \{0,1\} \times \mathcal{S} \rightarrow \mathbb{R}$ is a loss function appropriate for survival analysis. Common choices include the \textit{Brier score} for calibration, the \textit{negative log-likelihood}, and the \textit{C-index} for discrimination.

In the following subsections, we describe a methodology for quantifying the tunability of survival models. First, we define a dataset-level tunability metric that captures the gain in predictive performance achieved through hyperparameter tuning compared to default configurations. Then, we introduce a hyperparameter-level metric that isolates the contribution of each hyperparameter to the overall improvement. These measures enable both global and component-wise evaluation of tunability across datasets.

\subsection{Measuring Tunability of the Model}

A dataset-level tunability metric can be computed as the difference between the risk of a reference configuration (e.g., software defaults) and that of the best-performing configuration found through tuning:

\begin{equation}
d^{(j)} = R^{(j)}(\theta^*) - R^{(j)}(\theta^{(j)*}), \quad \text{for} \quad j = 1, \dots, m.
\end{equation}

\noindent Here, $R^{(j)}(\theta)$ denotes the risk (e.g., \textit{Brier score} or $1 -$ \textit{C-index}) of a model trained with hyperparameter setting $\theta$ on dataset $j$. $\theta^*$ is the default configuration, while $\theta^{(j)*}$ is the configuration that minimizes risk on dataset $j$ through tuning. The resulting difference, $d^{(j)}$, quantifies the performance gain due to tuning. These values across datasets can be summarized via mean, median, or quantiles to yield an overall tunability score $d$.

\subsection{Measuring Tunability of a Hyperparameter}

To assess the contribution of a specific hyperparameter, let $\theta_i^{(j)*}$ be the configuration where only the $i$-th hyperparameter is tuned on dataset $j$, while all others are kept at default values:

\begin{equation}
\theta_i^{(j)*} = \arg \min_{\theta \in \Theta, \, \theta_l = \theta_l^* \, \forall l \neq i} R^{(j)}(\theta).
\end{equation}

\noindent The tunability of hyperparameter $i$ on dataset $j$ is then measured by:

\begin{multline}
d^{(j)}_i = R^{(j)}(\theta^*) - R^{(j)}(\theta_i^{(j)*}), \\
\text{for} \quad j = 1, \dots, m, \quad i = 1, \dots, k.
\end{multline}

\noindent This value captures the performance improvement from tuning only parameter $i$, helping to identify which hyperparameters have the greatest individual impact. Additionally, we define the relative contribution as $d^{(j)}_{\text{rel}, i} = \frac{d^{(j)}_i}{d^{(j)}}$, which expresses the gain from tuning parameter $i$ as a fraction of the total gain. Aggregating $d^{(j)}_i$ over all datasets (e.g., via mean or quantiles) yields the overall tunability score $d_i$ for parameter $i$.

\section{Experiments}
In this paper, we investigate the tunability of the hyperparameters of the RSF model on the predictive maintenance task. The experiments are designed to explore the tunability of the RSF model, its hyperparameters, and optimal ranges. The materials for reproducing the experiments and the datasets are in the repository: \url{https://github.com/mcavs/RSF-tunability-paper}. 

\subsection{Modeling}

In the modeling phase, we trained RSF models using six hyperparameters, as detailed in Table~\ref{tab:rsf_hyperparameters}. The hyperparameter combinations were generated by exhaustively exploring a Cartesian product of multiple values for each hyperparameter, including different configurations such as $10$ for \texttt{ntree}, $10$ for \texttt{mtry}, $10$ for \texttt{nodesize}, 10 for \texttt{nodedepth}, $3$ for \texttt{splitrule}, and $11$ for \texttt{nsplit}—within predefined tuning ranges. This resulted in $330.000$ unique configurations for each dataset. This extensive search, integrated into a $5$-fold cross-validation framework, was designed to robustly validate model performance using the \textit{C-index} and the \textit{Brier score}.

\begin{table*}[h!]
\centering
\begin{threeparttable}
    \caption{\textbf{Hyperparameters of the RSF model}}
    \label{tab:rsf_hyperparameters}
    \setlength{\tabcolsep}{5pt}
    \renewcommand{\arraystretch}{1.2}
    \begin{tabular}{p{2cm} p{4.5cm} p{1.5cm} p{2.5cm} p{1.5cm} p{2.5cm}}
        \toprule
        \textbf{Hyperparameter} & \textbf{Description} & \textbf{Type} & \textbf{Range\textsuperscript{*}} & \textbf{Default} & \textbf{Tuned Range\textsuperscript{**}} \\
        \midrule
        \texttt{ntree}      & Number of trees in the forest                         & integer  & $[1, \infty)$                     & 500     & [100, 2000] \\
        \texttt{mtry}       & Number of variables randomly selected at each split   & numeric  & $[1, \infty)$                     & -       & [1, 10] \\
        \texttt{nodesize}   & Minimum number of observations in a terminal node          & integer  & $[1, \infty)$                     & 15      & [5, 100] \\
        \texttt{nodedepth}  & Maximum depth of each tree                            & integer  & $[1, \infty)$                     & -       & [5, 100] \\
        \texttt{splitrule}  & Criterion for splitting nodes                         & discrete & \{logrank, bs.gradient, logrankscore\} & logrank & \{logrank, logrankscore, bs.gradient\} \\
        \texttt{nsplit}     & Number of random splits per variable                  & integer  & $[0, \infty)$                     & 10      & [5, 15] \\
        \bottomrule
    \end{tabular}
    \begin{tablenotes}
    \small
    \item \textsuperscript{*}range of the hyperparameter, \textsuperscript{**}search range used in this paper
    \end{tablenotes}
\end{threeparttable}
\end{table*}            

\subsection{Datasets}

The CMAPSS dataset \citep{saxena_et_al_2008}, developed by NASA, simulates jet engine performance under various operational conditions. It is widely used in modeling studies due to its rich sensor measurements. The dataset consists of four subsets, each with distinct operational conditions and failure modes given in Table~\ref{tab:CMAPSS_summary}. \texttt{FD001} and \texttt{FD003} have a single operational condition, whereas \texttt{FD002} and \texttt{FD004} operate under six different conditions. Additionally, \texttt{FD003} and \texttt{FD004} include multiple failure modes. It is well-suited for evaluating the tunability potential of the RSF model for predictive maintenance strategies and identifying optimal configurations with its diverse operational conditions.

\begin{table}[H]
\centering
\caption{Summary of CMAPSS Dataset}
\label{tab:CMAPSS_summary}
\begin{tabular}{l c c c}
\toprule
\textbf{Subset} & \textbf{Sample Size} & \textbf{Condition} & \textbf{Fault Mode} \\ 
\midrule
\texttt{FD001} & 100  & 1 (Sea Level)           & HPCD     \\ 
\texttt{FD002} & 260  & 6 Different Conditions   & HPCD     \\ 
\texttt{FD003} & 100  & 1 (Sea Level)           & HPCD, FD \\ 
\texttt{FD004} & 248  & 6 Different Conditions   & HPCD, FD \\ 
\bottomrule
\end{tabular}
\vspace{0.1cm}

\scriptsize HPCD: \textit{High-Pressure Compressor Degradation}, FD: \textit{Fan Degradation}.
\end{table}

\section{Results}

To correctly interpret the results, it is important to note that higher values of the \textbf{\textit{C-index} indicate better model discrimination}, whereas lower values of the \textbf{\textit{Brier score} correspond to better calibration}. Accordingly, tunability metrics are computed to reflect improvements in model performance—i.e., increases in the \textit{C-index} and decreases in the \textit{Brier score}—when comparing tuned hyperparameters to default configurations. All reported tunability values are aligned with this convention, such that positive tunability indicates a beneficial change in the corresponding metric while a negative tunability value indicates a deterioration in model performance when the hyperparameter is tuned compared to the default.

\subsection{Tunability of the RSF Model}

The RSF model was trained separately on each of the four subsets of the CMAPSS dataset using both default and considered hyperparameter configurations. The results are presented in Table~\ref{tab:rsf_tunability_all}. 

\begin{table}[h!]
\centering
\caption{Tunability of RSF Models trained on the subsets of the CMAPSS datasets}
\label{tab:rsf_tunability_all}
\begin{tabular}{p{7mm}>{\centering\arraybackslash}p{7mm}>{\centering\arraybackslash}p{6mm}>{\centering\arraybackslash}p{12mm}>{\centering\arraybackslash}p{7mm}>{\centering\arraybackslash}p{6mm}>{\centering\arraybackslash}p{11mm}}

\toprule
& \multicolumn{3}{c}{\textbf{C-index}} & \multicolumn{3}{c}{\textbf{Brier score}}\\\cmidrule(lr){2-4}\cmidrule(lr){5-7}

\textbf{Subset} & \textbf{default} & \textbf{best} & \textbf{tunability} & \textbf{default} & \textbf{best} & \textbf{tunability} \\
\midrule
\texttt{FD001} & 0.7204 & 0.7705 & 0.0492 & 0.1422 & 0.1225 & 0.0197 \\
\texttt{FD002} & 0.6658 & 0.7294 & 0.0636 & 0.1484 & 0.1327 & 0.0157 \\
\texttt{FD003} & 0.8190 & 0.8718 & 0.0528 & 0.1158 & 0.0935 & 0.0222 \\
\texttt{FD004} & 0.7634 & 0.8169 & 0.0534 & 0.1309 & 0.1088 & 0.0221 \\
\midrule
\textbf{Average} &  &  & \textbf{0.0547} &  &  & \textbf{0.0199} \\
\bottomrule
\end{tabular}
\end{table}

As the table shows, hyperparameter tuning consistently improved both discrimination, measured by the \textit{C-index}, and calibration, reflected by the \textit{Brier score}. On average, the \textit{C-index} increased by $0.0547$, which corresponds to an approximate relative improvement of $7.5\%$ over the default configurations. Likewise, the \textit{Brier score} decreased by $0.0199$, which translates to an average relative reduction of $14.6\%$ in prediction error. These improvements were not only observed at the overall level but were also consistent across each subset. This demonstrates that the RSF model is sensitive to hyperparameter configurations and that tuning is not merely optional but should be regarded as necessary, particularly in predictive maintenance applications where survival prediction accuracy is essential.

\subsection{Tunability of the RSF Model's Hyperparameters}

\begin{figure}[h!]
    \centering
    \includegraphics[width = \linewidth]{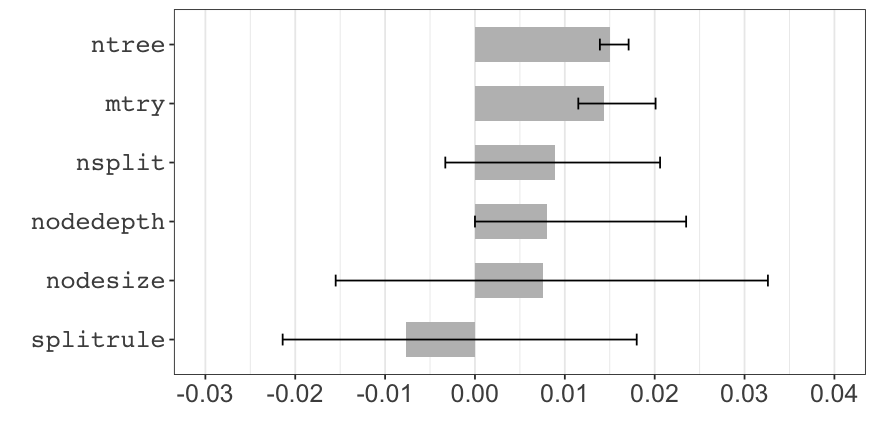}
    \caption{Tunability scores of hyperparameters in terms of \textit{C-index gain}. While the boxes show the mean tunability of each hyperparameter regarding the models on the four subsets, the error bars show the tunability range as minimum and maximum values.}
    \label{fig:tunability_plot}
\end{figure}

Figure~\ref{fig:tunability_plot} presents the average tunability scores of six hyperparameters used in the RSF model, based on the \textit{C-index}. Each bar represents the \textit{average C-index} improvement when a single hyperparameter is optimized while all others are kept at their default values, across four CMAPSS subsets. The error bars indicate the minimum and maximum tunability values observed within each dataset.

\texttt{ntree} and \texttt{mtry} hyperparameters exhibit the highest average tunability scores. This suggests that these two parameters play a more decisive role in improving model performance compared to the others. In contrast, the \texttt{nodesize}, \texttt{nodedepth}, and \texttt{nsplit} hyperparameters show lower but more stable average tunability values, implying a more limited yet consistent influence on the model. Notably, the \texttt{splitrule} hyperparameter has a negative average tunability score, indicating that if not carefully tuned, it may lead to decreased performance. These findings highlight that the impact of each hyperparameter varies when others are fixed at default settings, providing guidance on which parameters should be prioritized during the optimization process.

\begin{figure}[h!]
    \centering
    \includegraphics[width = \linewidth]{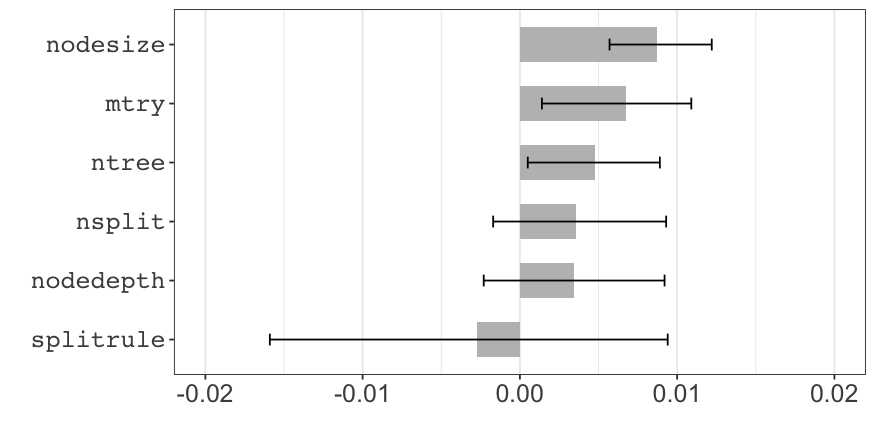}
    \caption{Tunability scores of hyperparameters in terms of \textit{Brier score gain}. While boxes show the mean tunability of each hyperparameter regarding the models on the four subsets, the error bars show the tunability range as minimum and maximum values.}
    \label{fig:figure_boxplot_brier}
\end{figure}

Figure~\ref{fig:figure_boxplot_brier} presents the average tunability scores of the hyperparameters used in the RSF model in terms of \textit{Brier Score}. Each bar represents the average improvement in \textit{Brier Score} obtained from the four CMAPSS subsets when one hyperparameter is optimized while the others are kept fixed at their default values. The error bars indicate the minimum and maximum tunability values observed.

Considering these results, the \texttt{nodesize} and \texttt{mtry} hyperparameters yield the greatest average reduction in \textit{Brier Score}. This indicates that these two parameters stand out from the others in their ability to positively influence model calibration during the analysis process.

When examining the other hyperparameters, \texttt{ntree}, \texttt{nsplit}, and \texttt{nodedepth} show lower tunability values. This suggests that although these parameters do not have a major impact on model performance, their influence is limited and consistent.

Another noteworthy finding relates to the \texttt{splitrule} hyperparameter. Its average tunability score hovers around zero and is accompanied by wide error margins. This indicates that while the \texttt{splitrule} hyperparameter may be beneficial in some datasets, it can be detrimental in others, and its impact should be considered carefully. These findings provide valuable insights into determining which hyperparameters should be prioritized during optimization to enhance model calibration.

\subsection{Tunability ranges of the RSF Model's Hyperparameters}

\begin{figure}[h!]
    \centering
    \includegraphics[width=\linewidth]{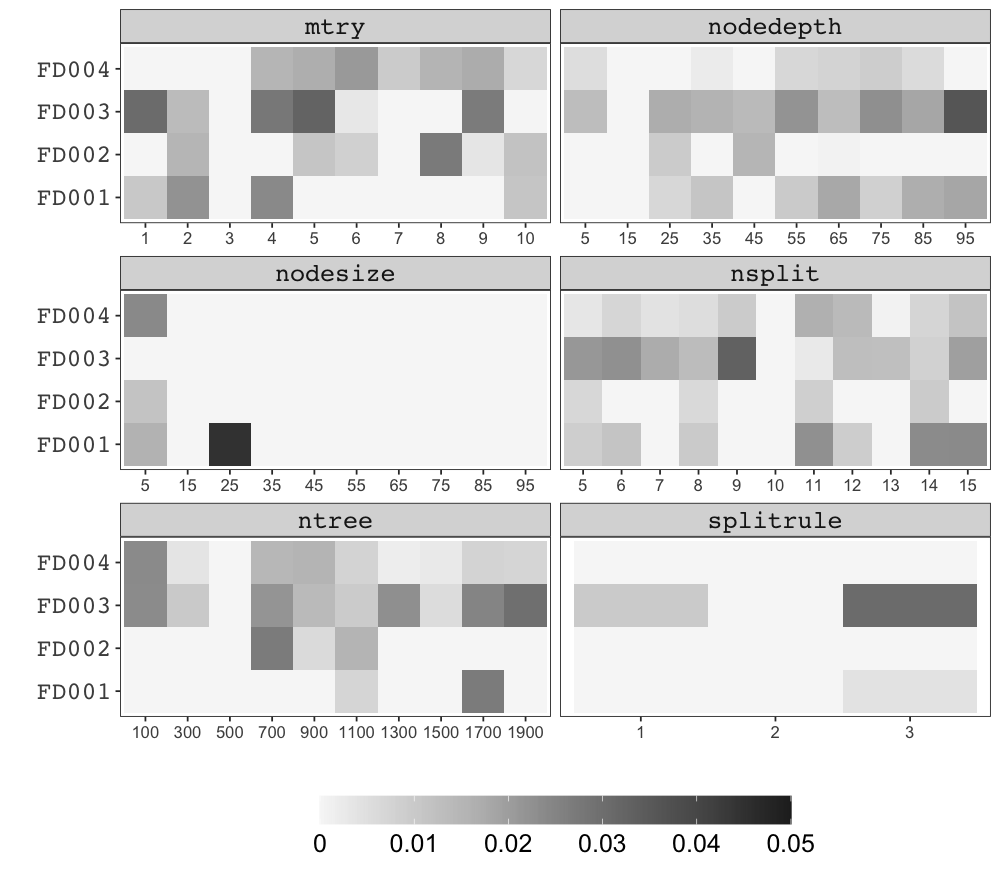}
    \caption{Tunability ranges of the hyperparameters in terms of \textit{C-index gain} across datasets and the hyperparameter values. The darker is better. The values of \texttt{splitrule} are coded as $1$: \texttt{bs.gradient}, $2$: \texttt{logrank}, and $3$: \texttt{logrankscore}.}
    \label{fig:hyperparam_grid}
\end{figure}

Figure~\ref{fig:hyperparam_grid} illustrates the effect of each hyperparameter on the \textit{C-index} performance of the RSF model across different value ranges. The tunability score corresponding to each hyperparameter value can be interpreted through color intensity. For example, lighter shades represent low tunability, while darker shades indicate high tunability.

According to the findings, some hyperparameters exhibit different patterns of influence depending on the range of values, and their effect tends to concentrate within specific intervals. For instance, the \texttt{nodesize} hyperparameter provides high performance gains at lower values (between $5$ and $15$), especially in the \texttt{FD001} and \texttt{FD004} datasets. This suggests that as the minimum number of observations in a terminal node decreases, the model can make more accurate decisions. Additionally, when the \texttt{nodesize} hyperparameter takes the value of $25$, it shows a noticeable impact on the \texttt{FD001} dataset. As for the \texttt{splitrule} hyperparameter, it appears to be meaningful only for \texttt{FD001} and \texttt{FD003}, and at specific values \texttt{bs.gradient} and \texttt{logrankscore}. This indicates that tuning this hyperparameter requires attention to those specific values.

When the other hyperparameters are examined, they exhibit broader and more scattered patterns of influence. Although some values of these hyperparameters improve performance in specific datasets, they are generally effective to varying degrees. In particular, there is no dominant value range for \texttt{mtry} and \texttt{ntree}; their impact varies depending on both the dataset and the modeling context. Overall, this visual analysis reveals that while some hyperparameters have a localized and distinct impact on model performance, others show more complex, dispersed, and dataset-specific variations.

\begin{figure}[h!]
    \centering
    \includegraphics[width=\linewidth]{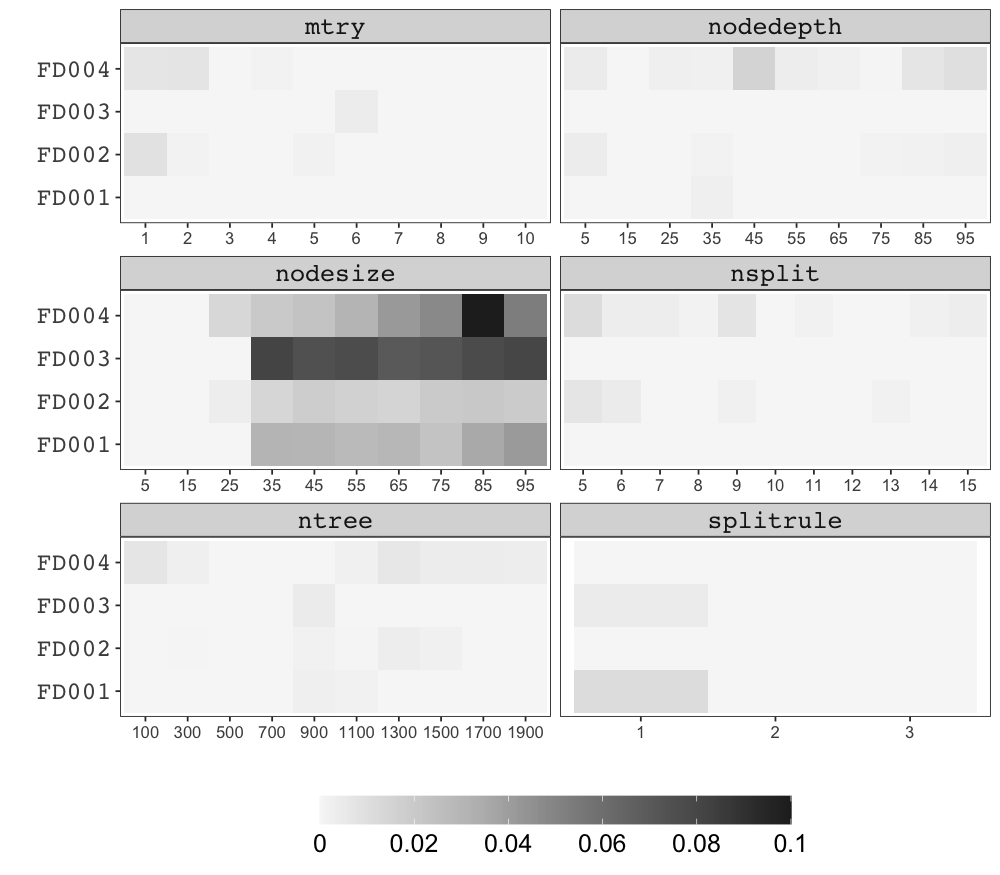}
    \caption{Tunability ranges of the hyperparameters in terms of \textit{Brier score gain} across datasets and the hyperparameter values. The darker is better. The values of \texttt{splitrule} are coded as $1$: \texttt{bs.gradient}, $2$: \texttt{logrank}, and $3$: \texttt{logrankscore}.}
    \label{fig:figure_heatmap_brier}
\end{figure}

Figure~\ref{fig:figure_heatmap_brier} shows the effect of each hyperparameter on the \textit{Brier score} performance of the RSF model across different values. The tunability corresponding to each hyperparameter value can be interpreted through color intensity.

\texttt{nodesize} stands out among all hyperparameters with a consistent and distinct pattern. Especially in the value range between $35$ and $95$, \texttt{nodesize} shows significantly darker areas in the \texttt{FD003} and \texttt{FD004} datasets. This suggests that a larger number of observations in a terminal node improves the calibration of the model. This finding is in contrast to the \textit{C-index} results, where a smaller number of observations in a terminal node was more beneficial. In this case, regarding the \textit{Brier score}, a larger number of observations in a terminal node appears to have a stronger effect.

The other hyperparameters generally display scattered and lighter-toned patterns. No strong or consistent value range significantly improves the \textit{Brier score}. According to the results obtained so far, \texttt{nodesize} emerges as a highly important hyperparameter for improving the \textit{Brier score}. In contrast, the other hyperparameters have more uncertain and unstable effects in terms of model calibration.

\section{Conclusions}

This study explored the tunability of the RSF model within the domain of predictive maintenance, highlighting the pivotal role of hyperparameter optimization in enhancing survival prediction performance. Leveraging comprehensive experiments on four distinct subsets of the CMAPSS dataset, we found that tuning RSF hyperparameters leads to meaningful improvements in both the model’s discriminatory power and its calibration quality. Specifically, hyperparameter tuning produced an average gain of $0.0547$ in the \textit{C-index}, corresponding to a relative improvement of approximately $7.5\%$ over the baseline configuration. In parallel, the \textit{Brier score} saw a mean reduction of $0.0199$, which translates to a relative decrease of about $14.6\%$ in prediction error. These enhancements were consistently observed across all dataset variations, underscoring that even modest absolute changes in these metrics can represent substantial relative improvements.

Among the hyperparameters examined, \texttt{ntree} and \texttt{mtry} emerged as the most influential in improving \textit{C-index}, suggesting their central role in enhancing the model’s ability to distinguish between failure times. In contrast, \texttt{nodesize} and \texttt{mtry} showed stronger and more stable effects on model calibration as measured by the \textit{Brier score}. \citet{hazewinkel2022prediction} reported that RSF predictions are highly sensitive to tuning, especially to the \texttt{nodesize} and \texttt{nodedepth} in clinical trials, which aligns with our findings of localized performance improvements for \texttt{nodesize}. Notably, \texttt{splitrule} exhibited negative average tunability, indicating that it may harm performance if not tuned carefully. Furthermore, visual analyses revealed that certain hyperparameters, such as \texttt{nodesize}, have localized regions (e.g., values between $10$ and $30$ for \textit{C-index}, or $35$ and $95$ for \textit{Brier score}) where performance gains are more pronounced, whereas others like \texttt{ntree} and \texttt{mtry} demonstrated more dataset-specific and scattered effects. On the other hand, \citet{dauda2022optimal} found that tuning only two core hyperparameters \texttt{mtry} and \texttt{nodesize} significantly enhanced RSF performance on a medical dataset, outperforming untuned RSF. Collectively, these studies provide empirical justification for our systematic evaluation of RSF tunability and demonstrate that careful hyperparameter optimization can be a decisive factor in real-world applications.

Despite its promising findings, this study has some limitations. First, the analysis was restricted to the CMAPSS dataset, which, although diverse in operational conditions and fault modes, may not fully represent other domains where survival modeling is applicable. As a result, the generalizability of tunability results to different industries or use cases remains uncertain. Second, the hyperparameter search space was predefined and limited to a fixed grid; this approach may overlook better configurations that lie outside the chosen ranges or that could be discovered through more advanced search strategies such as Bayesian optimization or evolutionary algorithms. Finally, model performance was evaluated solely based on two metrics without considering other aspects such as interpretability, computational cost, or temporal performance degradation.

Future work can expand upon these findings in several directions. One potential avenue is to evaluate RSF tunability on a broader range of datasets, particularly those from different domains with varying levels of censoring and data sparsity. Additionally, integrating automated hyperparameter optimization methods or meta-learning approaches could enable more efficient and adaptive tuning strategies. Another promising direction involves extending the notion of tunability beyond predictive performance to predictive multiplicity \citep{cavus2025}, explainability \citep{pashami2023explainable,cummins2024explainable}, and especially robustness, following the direction of the work in \citet{yardimci_et_al_2025}. Ultimately, this paper underscores the critical role of hyperparameter tuning in survival modeling and offers actionable insights into how practitioners can better exploit the RSF model’s capacity through informed tuning.

\section*{Acknowledgment}

The work on this paper is supported by the Artificial Intelligence Talent Cluster of Defence Industry Academic Thesis Program (SAYZEK - ATP), organized in collaboration with the Presidency of the Republic of Turkiye - Secretariat of Defense Industries and the Council of Higher Education.


\bibliographystyle{IEEEtran} 

\end{document}